\documentclass[sigplan]{acmart}

\usepackage{booktabs}
\usepackage{multirow}
\usepackage{xcolor}
\usepackage{enumitem}
\usepackage{graphicx}
\usepackage{tikz}
\usetikzlibrary{positioning,fit,backgrounds}

\setcopyright{none}
\renewcommand\footnotetextcopyrightpermission[1]{}
\acmConference[Agent Skills '26]{The First Workshop on Agent Skills: Design, Evaluation, and Optimization of Procedural Knowledge for LLM Agents}{May 26, 2026}{San Jose, CA, USA}
\acmYear{2026}
\copyrightyear{2026}

\begin{document}

\title{What Keeps Agent Skills from Being Reusable? Evidence from 138K SKILL.md Files}

\author{Chi Zhang}
\affiliation{%
  \institution{The Graduate Center, City University of New York}
  \city{New York}
  \state{NY}
  \country{USA}
}
\email{CHI.ZHANG06@gc.cuny.edu}
\author{Yimin Liu}
\affiliation{%
  \institution{The Ohio State University}
  \city{Columbus}
  \state{OH}
  \country{USA}
}
\email{yiminliu.career@gmail.com}

\author{Xinze Chen}
\affiliation{%
  \institution{The Graduate Center, City University of New York}
  \city{New York}
  \state{NY}
  \country{USA}
}
\email{xinze.chen95@gc.cuny.edu}

\author{Ping Ji}
\affiliation{%
  \institution{Hunter College, City University of New York}
  \city{New York}
  \state{NY}
  \country{USA}
}
\email{ping.Ji@hunter.cuny.edu}

\begin{abstract}
Under the current standard, Agent Skills are \texttt{SKILL.md} files that combine instructions with supporting resources, enabling Large Language Model (LLM) agents to reuse procedures beyond a single conversation.
Yet many public skills appear to originate from a single task, repository, or conversation, even when they are shared as reusable components.
We analyze this gap across 138,133 public \texttt{SKILL.md} files from 20,556 repositories using a two-tier defect taxonomy grounded in the official specification and best-practice guidance.
We find that 91.8\% of skills contain at least one detected defect, with stable estimates across lenient and strict thresholds (88.8--94.6\%).
The dominant failures are ordinary packaging problems rather than exotic attacks: weak routing metadata, bloated or non-actionable bodies, and poor resource organization.
A deterministic routing stress test over 20,000 skills shows the functional impact: skills with valid routing metadata are retrieved more reliably from startup descriptions than skills with routing defects.
Defect rates vary by platform and provenance: specification-aware skills contain fewer defects, while AI-marked skills show more safety and portability problems.
Lightweight enforcement and repair experiments support a quality-assured generation workflow combining spec-aware prompting, lightweight linting, automated repair, and safety gating.
\end{abstract}

\keywords{Agent Skills, LLM Agents, SKILL.md, Reusability Defects, Skill Routing, Quality-Assured Generation}

\maketitle

\section{Introduction}
\label{sec:intro}

Recently, AI skills have become a practical way to extend Large Language Model (LLM) agents beyond the immediate prompt~\cite{greshake2023indirect}.
Under the Agent Skills standard, a skill is packaged as a \texttt{SKILL.md} file: frontmatter tells the agent when to load it, the body gives instructions, and optional files provide scripts, references, or assets~\cite{agentskills-spec}.
In this form, skills turn task experience into reusable software artifacts.

That promise depends on a basic assumption: a published skill should still work outside the task, repository, platform, or conversation that produced it.
Our crawl contains 138,133 public \texttt{SKILL.md} files from 20,556 repositories, and its growth has outpaced quality control.
Many skills appear to be written for a narrow local purpose and then shared as reusable components.
Some omit official routing or structure requirements; others inline too much code, include setup notes or changelogs, leak local paths or credentials, or depend on one model, platform, or operating system.
These are not just formatting issues.
For a skill to be reusable, an agent must be able to select it, load it, locate supporting resources, execute it safely, and port it beyond the author's original environment.
Each stage corresponds to one or more categories in our taxonomy, distinguishing our defects from generic style or quality issues.

Existing work captures parts of this problem.
SkillsBench~\cite{li2026skillsbenchbenchmarkingagentskills} shows that curated skills improve task completion by 16.2 percentage points while self-generated skills provide no measurable benefit.
SkillReducer~\cite{gao2026skillreduceroptimizingllmagent} shows that non-actionable content wastes context, Wang et al.~\cite{wang2026elementarydearwatsondetecting} find malicious skills in public registries, and Liu et al.~\cite{liu2026agenticskillsworkwild} show that realistic skill retrieval remains brittle.
These studies make clear that skill quality matters, but they leave a more basic question open: when a public \texttt{SKILL.md} file fails to transfer across tasks or platforms, what exactly has gone wrong?
We answer this question by characterizing whether public skills have the artifact properties required for reuse.

We organize the study around four questions:
\begin{itemize}[leftmargin=*]
    \item \textbf{RQ1:} What reusability defects are prevalent in public skills?
    \item \textbf{RQ2:} How do these defects limit skill reuse?
    \item \textbf{RQ3:} Which platform and provenance signals are associated with reusable skill quality?
    \item \textbf{RQ4:} What traits characterize reusable, high-quality Agent Skills?
\end{itemize}

We make four contributions:
\begin{enumerate}[leftmargin=*]
    \item \textbf{A reusable-skill quality model:} a two-tier taxonomy of 7 categories and 31 checks grounded in official requirements and best-practice guidance.
    \item \textbf{An ecosystem-scale diagnosis:} an analysis of 138,133 public skills showing that 89.3\% violate the official specification and 91.8\% contain at least one detected defect.
    \item \textbf{Functional evidence that defects matter:} a routing stress test showing that R1-clean skills are retrieved more reliably than R1-defective skills from startup descriptions.
    \item \textbf{A path from diagnosis to intervention:} platform, provenance, exemplar, enforcement, and repair analyses that motivate a quality-assured generation workflow and twelve evidence-based authoring guidelines.
\end{enumerate}

The remainder of the paper is organized as follows.
Section~\ref{sec:background} reviews the Agent Skills format and related work.
Section~\ref{sec:method} describes our dataset, defect taxonomy, detector validation, and routing stress test.
Section~\ref{sec:results} answers the four research questions.
Sections~\ref{sec:discussion} and~\ref{sec:threats} discuss implications and limitations, and Section~\ref{sec:conclusion} concludes.
\section{Background and Related Work}
\label{sec:background}

\subsection{Agent Skills and the SKILL.md Format}

An Agent Skill is a directory containing a \texttt{SKILL.md} file with two parts: YAML (Yet Another Markup Language) frontmatter for structured metadata and a Markdown body for instructions.
The frontmatter must include a \texttt{name} (max 64 characters, lowercase alphanumeric and hyphens) and a \texttt{description} (max 1,024 characters) that specifies \emph{what} the skill does and \emph{when} to use it~\cite{agentskills-spec}.
The description serves as the primary routing mechanism: agents load skill descriptions at startup to decide which skill to activate for a given task.
The body contains instructions the agent follows when the skill is invoked.

The specification recommends keeping the body under 500 lines and organizing supporting content into \texttt{scripts/}, \texttt{references/}, and \texttt{assets/} subdirectories~\cite{agentskills-spec}.
This \emph{progressive disclosure} pattern loads metadata ($\sim$100 tokens) at startup, the body ($<$5,000 tokens) on activation, and resources only when needed, minimizing context window consumption.

\subsection{Related Work}

The recent literature has largely established that skills can help agents, but only under the right conditions.
SkillsBench~\cite{li2026skillsbenchbenchmarkingagentskills} makes this contrast especially clear: curated skills improve task completion by 16.2 percentage points, while self-generated skills provide \emph{no measurable benefit}.
The same study finds that focused skills with 2--3 modules outperform comprehensive documentation, suggesting that a useful skill is not simply a longer prompt or a larger knowledge dump.
Liu et al.~\cite{liu2026agenticskillsworkwild} reach a similar conclusion from a different angle: when agents must retrieve from 34K real-world skills, the benefit of skills becomes brittle.
These findings move the question upstream.
Before asking whether a skill improves a task, we need to ask whether the skill is written in a form that can be selected, loaded, and reused outside the context where it was created.

That upstream question turns skills into an artifact-quality problem.
A \texttt{SKILL.md} file sits between prompt engineering, agent scaffolding, and software packaging: it must be concise enough for progressive disclosure, precise enough for routing, and concrete enough to guide execution.
SkillReducer~\cite{gao2026skillreduceroptimizingllmagent} studies one part of this artifact problem by showing that over 60\% of body content in public skills is non-actionable and that 26.4\% of skills lack routing descriptions.
Those results explain why token efficiency and routing metadata matter, but they also point beyond compression.
Public skills can fail through body bloat, weak descriptions, poor resource organization, local environment assumptions, unsafe commands, or conflicts with the agent's surrounding instructions.
Our work therefore treats skill quality as a multi-stage reusability question rather than a single optimization target.

Agent harnesses make this question more consequential.
ClawsBench~\cite{li2026clawsbenchevaluatingcapabilitysafety} evaluates LLM productivity agents across realistic Gmail, Slack, Calendar, Docs, and Drive workflows, where domain skills expose API knowledge through progressive disclosure and a meta prompt coordinates behavior across services.
In this setting, a skill is no longer a private note saved from one conversation.
It becomes part of the interface between the agent, the harness, and external tools.
For such an interface to work, a skill needs routable metadata, manageable instructions, organized supporting resources, and enough portability to survive changes in platform, model, or operating environment.

The same interface also raises safety concerns.
Wang et al.~\cite{wang2026elementarydearwatsondetecting} analyze 150,108 skills from 7 registries and find 620 malicious skills, while Skill-Inject~\cite{schmotz2026skillinjectmeasuringagentvulnerability} shows that injected skill files can reach up to 80\% attack success.
These works study deliberately malicious skills, but their threat model highlights a broader issue for public reuse: once a skill is shared, the agent may treat its contents as operational instructions.
Even benign skills can therefore be risky if they preserve hardcoded credentials, unsafe shell commands, local paths, or instructions that redefine the agent's role.
This connects skill quality to general LLM-agent safety concerns, including OWASP risks such as prompt injection and excessive agency~\cite{owasp2025}, as well as instruction-hierarchy failures where lower-priority instructions can still override higher-priority intent~\cite{wallace2024instructionhierarchytrainingllms,geng2025control,guo2026ihchallengetrainingdatasetimprove}.

Tooling has started to address parts of this problem.
The \texttt{agnix} linter~\cite{agnix} implements 385 validation rules across 6 agent platforms, and the official \texttt{skills-ref} validator~\cite{agentskills-spec} checks frontmatter syntax.
What remains missing is an ecosystem-level account of which defects are actually common, which ones affect functional reuse, and which checks should be enforced before repair or human review.
Our study fills this gap by characterizing 138,133 public \texttt{SKILL.md} files with a two-tier taxonomy that spans routing, body design, resource organization, prohibited content, behavioral safety, portability, and persona/scope conflicts.

\section{Methodology}
\label{sec:method}

\subsection{Data Collection}
\label{sec:collection}

To study reuse rather than isolated benchmark performance, we build the dataset from places where skills are actually published and copied.
We collected \texttt{SKILL.md} files through GitHub code search, repository cloning, and the agentskills.in registry API.
GitHub search used 40+ sharded queries to bypass the 1,000-result-per-query limit; repository cloning extracted all case-insensitive \texttt{SKILL.md} files from known and discovered skill repositories; registry collection fetched raw GitHub content for indexed skills.

The final SHA-256-deduplicated dataset contains 138,133 unique skills from 20,556 repositories; 98.8\% have YAML frontmatter, 98.5\% include a \texttt{name}, 98.1\% include a \texttt{description}, and the median body length is 169 lines / 687 words.

\paragraph{Inclusion criteria and representativeness.}
We applied minimal filtering to preserve an ecosystem-wide view rather than a curated subset.
The only filters were (i) the file must be named \texttt{SKILL.md} (case-insensitive), (ii) the host repository must be public at crawl time, and (iii) we keep one copy per SHA-256 content hash to remove verbatim duplicates.
We did not filter by repository popularity (stars), freshness (last commit date), or author identity, because two questions in this paper---ecosystem-wide prevalence and platform/provenance variation---require including unpopular and stale repositories rather than excluding them.
For context, repository-level skew is non-trivial: the most prolific repository contributes 17,284 skills (12.9\%) and 47.0\% of repositories contribute exactly one skill, so we report both aggregate and per-repository statistics throughout, and we use repository-stratified subsamples for the routing stress test (Section~\ref{sec:routing-stress}) and the exemplar analysis (RQ4) to avoid letting one repository dominate.
A small enterprise or private-repo bias is unavoidable; we address its external-validity implications in Section~\ref{sec:threats}.

We also collected GitHub issues from 12 major agent platform repositories using 28 skill-related queries; filtering 6,233 issues yielded 761 relevant issues classified by defect category.

\subsection{Defect Taxonomy}
\label{sec:taxonomy}

We organize defects around the lifecycle of reuse.
A skill must first be selected, then loaded, then supported by external resources, then executed safely in an environment that may differ from the one where it was authored.
We developed the taxonomy through three iterative passes.
First, we read the official Agent Skills specification~\cite{agentskills-spec} and the Claude Code skills documentation~\cite{claudecode-skills} clause by clause, extracting each verifiable structural requirement into a candidate Tier~1 detector.
Second, we coded a stratified sample of 300 skills (drawn across platforms and size buckets) for failure patterns that did not map onto a spec clause but recurred in practice---install boilerplate, hardcoded credentials, persona redefinition, platform-specific paths---and clustered them against the peer-reviewed literature on prompt injection, instruction hierarchy, hardcoded secrets, and software portability~\cite{wallace2024instructionhierarchytrainingllms,geng2025control,cwe798,iso25010} to form Tier~2.
Third, we cross-validated the candidate categories against the 761 GitHub issues we collected (Section~\ref{sec:collection}): every category in the final taxonomy is grounded in either a spec clause or an observed real-world failure mode reported by skill consumers.
The two tiers are kept separate by design so that compliance violations and best-practice issues can be reported and prioritized independently rather than blurred into a single ``defect'' bucket.
We use the term \emph{detected defect} for a measurable property of a skill file that either violates the official specification or contravenes established best practices, and for which we can point to documented evidence of real-world harm (GitHub issues, benchmark regressions, or security advisories).
The taxonomy is designed for static characterization: it identifies skills that may be harder to route, load, maintain, execute safely, or reuse outside their original context.
Each rule was operationalized as a regex-based or structural detector applied to the parsed skill.

\paragraph{Tier 1: Spec Conformance (14 checks).}
These rules are grounded directly in the official Agent Skills specification~\cite{agentskills-spec} and Claude Code documentation~\cite{claudecode-skills}.
Violations affect the core reuse path: missing or non-functional descriptions cause routing failures, oversized bodies increase context cost, and redundant content wastes scarce context tokens.
\begin{itemize}[leftmargin=*]
    \item \textbf{R1. Routing Metadata} (6 checks): missing, too-short, too-long, or non-functional descriptions; routing information misplaced in body.
    \item \textbf{R2. Body Content} (5 checks): oversized bodies, non-actionable content, obvious explanations, name-as-heading duplication, description duplication.
    \item \textbf{R3. Resource Organization} (3 checks): excessive inline code, too many examples, monolithic skills (i.e., large skills that do not use the recommended \texttt{scripts/}/\texttt{references/} subdirectories).
\end{itemize}

\paragraph{Tier 2: Best Practice Compliance (17 checks).}
These rules are derived from peer-reviewed security and software engineering literature.
\begin{itemize}[leftmargin=*]
    \item \textbf{R4. Prohibited Content} (4 checks): install instructions, changelogs, license text, unfinished markers (e.g., \texttt{TODO}, \texttt{FIXME})~\cite{gao2026skillreduceroptimizingllmagent}.
    \item \textbf{R5. Behavioral Safety} (6 checks): hardcoded credentials, safety bypasses (e.g., \texttt{--no-verify}), prompt injection patterns, unguarded destructive actions (e.g., \texttt{rm -rf}), error suppression, user path leakage~\cite{cwe798,owasp2025,wang2026elementarydearwatsondetecting}.
    \item \textbf{R6. Portability} (4 checks): hardcoded model names (e.g., \texttt{gpt-4o}), platform-specific paths, platform-specific tool calls, OS-specific commands (e.g., \texttt{pbcopy})~\cite{agentskills-spec,iso25010}.
    \item \textbf{R7. Persona \& Scope} (3 checks): persona redefinition (e.g., ``you are a\ldots''), instruction overrides, scope mismatches~\cite{wallace2024instructionhierarchytrainingllms,geng2025control}.
\end{itemize}

\begin{figure*}[htbp]
\centering
\resizebox{\textwidth}{!}{%
\begin{tikzpicture}[font=\sffamily,
    root/.style={draw, rounded corners, fill=black!85, text=white, font=\sffamily\bfseries\normalsize, minimum height=10mm, inner sep=6pt},
    tier/.style={draw, rounded corners, fill=gray!18, font=\sffamily\bfseries\small, minimum height=9mm, inner sep=5pt},
    cat/.style={draw, rounded corners, font=\sffamily\small\bfseries, minimum height=9mm, inner sep=4pt, text width=28mm, align=center},
    leaf/.style={draw, rounded corners=2pt, fill=white, font=\sffamily\small, text width=30mm, align=left, inner sep=4pt},
    spec/.style={cat, fill=blue!12},
    best/.style={cat, fill=orange!12},
    arr/.style={draw, ->, >=stealth, thick, gray!70},
]

\node[root] (root) at (0, 0) {Reusability Defects in Agent Skills (7 categories, 31 checks)};

\node[tier] (t1) at (-4.0, -1.5) {Tier 1: Spec Conformance};
\node[tier] (t2) at ( 4.0, -1.5) {Tier 2: Best Practice};

\draw[arr] (root) -- (t1);
\draw[arr] (root) -- (t2);

\node[spec] (r1) at (-8.8, -3.2) {R1. Routing\\Metadata\\{\scriptsize(6 checks)}};
\node[spec] (r2) at (-4.8, -3.8) {R2. Body\\Content\\{\scriptsize(5 checks)}};
\node[spec] (r3) at (-0.8, -3.2) {R3. Resource\\Organization\\{\scriptsize(3 checks)}};

\draw[arr] (t1) -- (r1);
\draw[arr] (t1) -- (r2);
\draw[arr] (t1) -- (r3);

\node[best] (r4) at (3.2, -3.2) {R4. Prohibited\\Content\\{\scriptsize(4 checks)}};
\node[best] (r5) at (6.6, -3.8) {R5. Behavioral\\Safety\\{\scriptsize(6 checks)}};
\node[best] (r6) at (10.0, -3.2) {R6. Portability\\{\scriptsize(4 checks)}};
\node[best] (r7) at (13.4, -3.8) {R7. Persona \&\\Scope\\{\scriptsize(3 checks)}};

\draw[arr] (t2) -- (r4);
\draw[arr] (t2) -- (r5);
\draw[arr] (t2) -- (r6);
\draw[arr] (t2) -- (r7);

\node[leaf] (r1l) at (-8.8, -5.9) {
R1.1 Missing description\\
R1.2 Too short ($<$30ch)\\
R1.3 Missing ``when''\\
R1.4 Non-functional\\
R1.5 Too long ($>$300ch)\\
R1.6 Routing misplacement
};

\node[leaf] (r2l) at (-4.8, -6.8) {
R2.1 Exceeds 500 lines\\
R2.2 Non-actionable body\\
R2.3 Explains obvious\\
R2.4 Name as heading\\
R2.5 Desc.\ dup.\ in body
};

\node[leaf] (r3l) at (-0.8, -5.6) {
R3.1 Excessive inline code\\
R3.2 Too many examples\\
R3.3 Monolithic (no refs)
};

\node[leaf] (r4l) at (3.2, -5.6) {
R4.1 Install instructions\\
R4.2 Changelog\\
R4.3 License text\\
R4.4 Unfinished markers
};

\node[leaf] (r5l) at (6.6, -6.8) {
R5.1 Hardcoded creds\\
R5.2 Bypass safety\\
R5.3 Prompt injection\\
R5.4 Unguarded actions\\
R5.5 Suppress errors\\
R5.6 Leak user paths
};

\node[leaf] (r6l) at (10.0, -5.6) {
R6.1 Hardcoded model\\
R6.2 Platform path\\
R6.3 Platform tools\\
R6.4 OS-specific cmds
};

\node[leaf] (r7l) at (13.4, -6.5) {
R7.1 Redefines persona\\
R7.2 Overrides prior instr.\\
R7.3 Scope mismatch
};

\draw[arr] (r1.south) -- (r1l.north);
\draw[arr] (r2.south) -- (r2l.north);
\draw[arr] (r3.south) -- (r3l.north);
\draw[arr] (r4.south) -- (r4l.north);
\draw[arr] (r5.south) -- (r5l.north);
\draw[arr] (r6.south) -- (r6l.north);
\draw[arr] (r7.south) -- (r7l.north);

\begin{scope}[on background layer]
    \node[draw=blue!60, dashed, very thick, rounded corners=8pt, fill=blue!3,
          fit=(t1)(r1)(r2)(r3)(r1l)(r2l)(r3l),
          inner xsep=12pt, inner ysep=8pt,
          label={[font=\footnotesize\itshape, text=blue!60]below:{Grounded in Official Agent Skills Spec}}] {};
\end{scope}

\end{tikzpicture}%
}
\caption{Two-tier defect taxonomy. Dashed box: Tier~1 (spec conformance). Tier~2: peer-reviewed literature and industry standards.}
\label{fig:taxonomy}
\end{figure*}
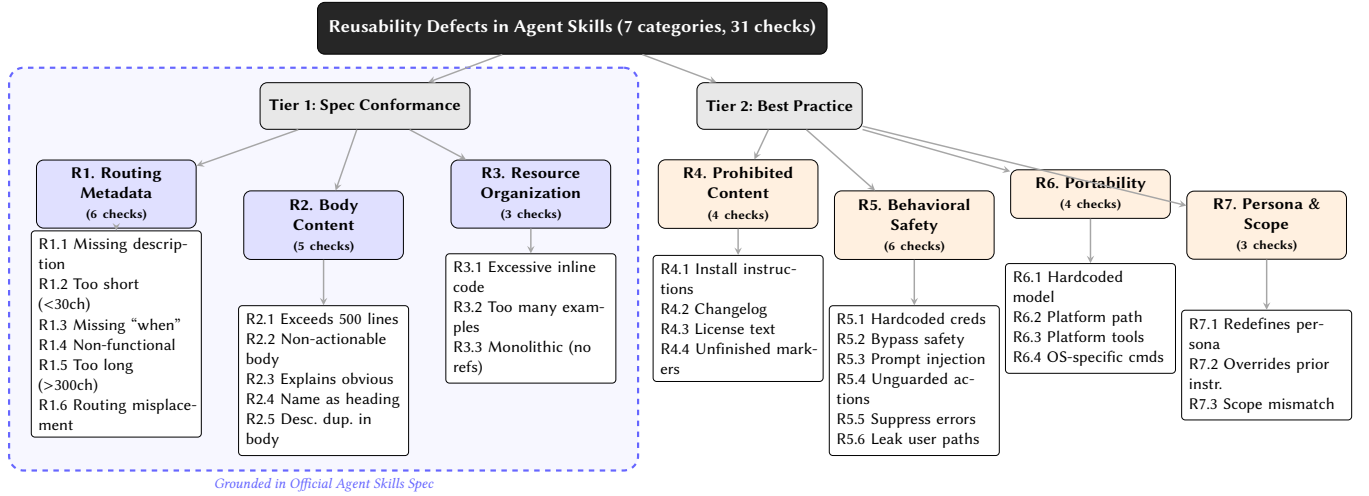

Figure~\ref{fig:taxonomy} presents the complete taxonomy as a hierarchical tree.
Tier~1 categories (dashed box) are grounded in the official specification; Tier~2 categories are grounded in peer-reviewed literature and industry standards.

The resulting taxonomy contains 31 checks across the seven categories listed above.

\subsection{Detector Validation}

We validate detectors with mutation-style \texttt{SKILL.md} fixtures and threshold sensitivity checks.
Each of the 31 detectors has a positive fixture that isolates the target defect and a paired negative fixture; all detectors fire only on the positive case.
For numeric thresholds (description length, body length, inline-code ratio, code-block count, monolithic-body length), lenient, baseline, and strict settings on a deterministic 10,000-skill sample preserve the headline result: 88.8--94.6\% of skills contain at least one detected defect, and the top three categories remain R1 Routing, R2 Body, and R3 Resources.

\subsection{Routing Stress Test}
\label{sec:routing-stress}

We also run a deterministic retrieval stress test over 20,000 skills.
We build a BM25 index over frontmatter descriptions only and ask whether metadata-derived or name/path-derived queries retrieve the source skill.
This is not an end-to-end agent benchmark, and we do not claim BM25 mirrors production routing: shipped harnesses such as Claude Code and Cursor use LLM-as-selector over the full description set rather than lexical scoring.
We choose lexical retrieval deliberately as a \emph{lower-bound probe}: it is the simplest mechanism for which a routing-defective description (missing, too short, missing trigger language, or non-functional) can plausibly degrade discovery.
A more semantic retriever (embedding reranker or LLM selector) may compress or shift the gap, particularly when descriptions are phrased differently than queries; we treat the BM25 result as evidence that routing-metadata defects have \emph{at least} the magnitude observed here at the discovery stage, and we flag the semantic-retrieval condition as future work in Section~\ref{sec:threats}.

\section{Results}
\label{sec:results}

\subsection{Ecosystem Overview}

Before examining quality, we first ask what kind of ecosystem these public files represent.
The answer matters because a reuse problem has different implications in a small curated registry than in a large, heterogeneous collection of repository-local artifacts.

We infer platforms from file paths.
Claude Code is the largest identifiable platform (18.7\%), while generic \texttt{skills/} directories dominate overall (58.3\%).
Skills primarily target development workflow automation: debugging (53.1\%), documentation (51.0\%), API development (48.0\%), testing (42.2\%), and code review (31.4\%).
Claude is mentioned in 25.4\% of skill content, far exceeding GPT-4/OpenAI (3.4\%), Gemini (3.2\%), and Copilot (1.4\%).
At least 14.4\% of skills carry explicit AI-generation markers, a conservative lower bound because many generated files are unmarked.
The ecosystem is also concentrated: 47.0\% of repositories contain one skill, while the largest contributor provides 17,284 skills (12.9\%).

\subsection{RQ1: What reusability defects are prevalent in public skills?}

\begin{figure}[htbp]
\centering
\includegraphics[width=\columnwidth]{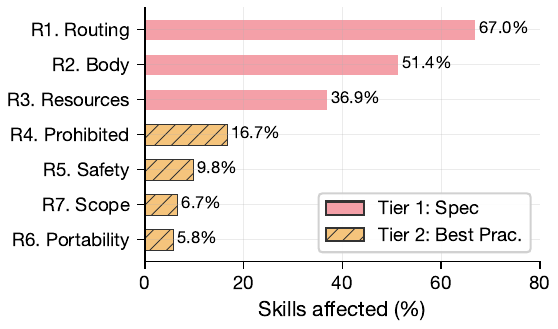}
\caption{Defect prevalence by category. Tier~1 (red, solid): spec conformance violations. Tier~2 (orange, hatched): best practice violations.}
\label{fig:prevalence}
\end{figure}

The first step is prevalence: if detected defects are rare, reusable-skill quality is mainly an edge-case problem; if they are common, quality control must become part of the skill generation and publication workflow.
Of 138,133 skills analyzed, 89.3\% trigger at least one Tier~1 detector derived from the official Agent Skills specification, and 91.8\% have at least one detected defect of any tier.
The average skill contains 2.5 detected defects (median: 2).
Figure~\ref{fig:prevalence} shows prevalence by category.
This prevalence result should be read as an ecosystem-level quality signal: some detected defects block routing or create safety risk, while others mostly waste context or indicate weak packaging.

\paragraph{Spec conformance defects dominate.}
The three most prevalent individual defects are all spec violations:
R1.3 missing trigger guidance (52.3\%), R2.4 name-as-heading duplication (44.3\%), and R3.2 too many inline examples (32.1\%).
67.0\% of skills have at least one routing defect; combined with R1.4 (13.5\%, non-functional descriptions), over half of all skills have descriptions that cannot effectively guide agent routing.

\subsection{RQ2: How do these defects limit skill reuse?}

\paragraph{Routing defects reduce discovery quality.}
Prevalence alone does not show whether a defect matters operationally.
The most direct reuse failure is that an otherwise useful skill is never selected.
Because R1 is both prevalent and tied to startup selection, we test this discovery-stage reuse risk in Table~\ref{tab:routingstress}.
When queries are derived from repository metadata, R1-clean skills achieve 88.5\% hit@1 and 0.906 MRR, compared with 82.6\% hit@1 and 0.855 MRR for R1-defective skills.
The same direction holds under the stricter name/path-only query proxy (26.7\% vs.\ 24.3\% hit@1).
\paragraph{What the stress test does and does not show.}
The stress test isolates one stage of reuse---discovery from the startup description surface---and one defect category, R1.
We chose R1 because it is the only category whose harms can be measured by a self-contained retrieval experiment over the artifact: a missing or non-functional description \emph{must} reduce discovery probability, and that effect can be quantified without an external task harness.
For R2--R3, the operational impact is context cost and progressive-disclosure friction; for R4--R6, it is content pollution, execution safety, and portability; for R7, it is conflict with the host agent's instruction hierarchy.
Each of these requires either an end-to-end agent benchmark or a deployment harness to measure faithfully, which is outside our scope here.
We address the remaining categories through three indirect lines of evidence in the rest of the paper: GitHub-issue corroboration (Section~\ref{sec:discussion}), the automated repair experiment for R5 (Section~\ref{sec:discussion}), and prior task-success benchmarks that already report sensitivity to body bloat and resource organization~\cite{li2026skillsbenchbenchmarkingagentskills,gao2026skillreduceroptimizingllmagent}.

\begin{table}[htbp]
\caption{Routing stress test over 20,000 sampled skills. The index contains frontmatter descriptions only.}
\label{tab:routingstress}
\centering
\small
\begin{tabular}{llrrr}
\toprule
Query & Group & $n$ & Hit@1 & MRR \\
\midrule
\multirow{2}{*}{Metadata}
& R1-clean & 6,514 & 88.5\% & 0.906 \\
& R1-defective & 13,226 & 82.6\% & 0.855 \\
\midrule
\multirow{2}{*}{Name/path}
& R1-clean & 6,505 & 26.7\% & 0.347 \\
& R1-defective & 13,331 & 24.3\% & 0.322 \\
\bottomrule
\end{tabular}
\end{table}

\subsection{RQ3: Which platform and provenance signals are associated with reusable skill quality?}

\begin{figure}[htbp]
\centering
\includegraphics[width=\columnwidth]{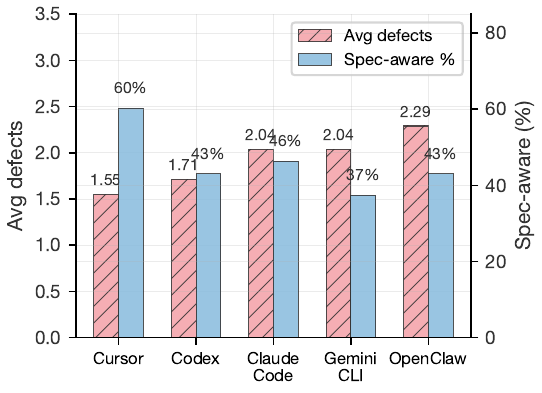}
\caption{Cross-platform quality: average defects (red) vs.\ specification awareness (blue).}
\label{fig:crossplatform}
\end{figure}

\paragraph{Cross-platform quality.}
If defects reflect only individual author mistakes, platform-level patterns should be weak.
Instead, the data suggest that platform conventions and provenance signals travel with different skill-quality profiles.
Figure~\ref{fig:crossplatform} reveals statistically significant quality variation across platforms (Kruskal--Wallis $H = 594.17$, $p < 10^{-125}$).
Cursor skills are highest quality (avg 1.55 defects, 13.5\% defect-free), while OpenClaw marketplace skills are lowest (avg 2.29, 3.1\% defect-free).
Cursor differs significantly from Generic, OpenClaw, and Claude Code (Bonferroni-corrected Mann--Whitney tests, all $p_{\mathrm{adj}} < 10^{-34}$).
Three factors correlate with platform-level quality: specification awareness (spec-aware skills average 1.83 defects vs.\ 3.00 for spec-unaware, Cliff's $\delta = -0.40$), AI-generation rate, and curation model.

\paragraph{AI-marked skills differ from unmarked skills.}

\begin{figure}[htbp]
\centering
\includegraphics[width=\columnwidth]{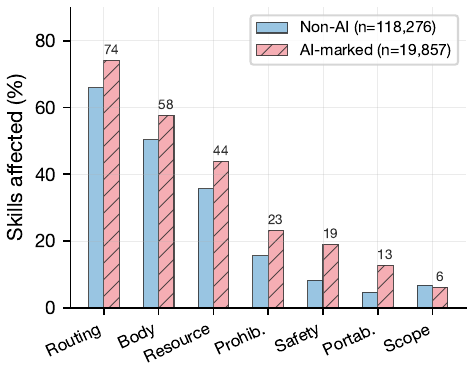}
\caption{Defect rates by category: AI-marked skills (red) vs.\ unmarked skills (blue). AI-marked skills show higher detected defect rates across all categories, with the largest gaps in Safety (2.3$\times$) and Portability (2.8$\times$).}
\label{fig:aiprovenance}
\end{figure}

Figure~\ref{fig:aiprovenance} compares defect rates between the 19,857 skills with explicit AI-generation markers and the remaining 118,276 unmarked skills.
We emphasize that this is an \emph{association} between an observable marker and observed defect rates, not a causal claim about AI vs.\ human authorship.
The AI-marked subset is not a random sample of AI-generated skills: only some generators or authors self-disclose, sloppier authors may be more likely to ship the boilerplate marker, and some platforms auto-stamp markers, all of which can produce a confounded comparison.
AI-marked skills average 3.23 detected defects versus 2.34 for unmarked skills, a 38.0\% increase (Mann--Whitney $U$, $p < 10^{-300}$; Cliff's $\delta = +0.28$, small effect), and are defect-free at half the rate (5.0\% vs.\ 8.8\%).
The gap is sharpest for safety-critical categories: 18.9\% of AI-marked skills have behavioral safety defects (R5), compared to 8.2\% of unmarked skills (2.3$\times$; $\chi^2 = 2{,}209$, $p < 10^{-300}$, $\phi = 0.13$).
Portability defects (R6) show a similar pattern (12.8\% vs.\ 4.6\%, 2.8$\times$), consistent with generated or conversation-derived skills inheriting platform-specific paths and model references.
We read this pattern as a signal that the explicit-marker subset warrants closer review at generation time, not as evidence that AI-authored skills are intrinsically worse; a propensity-matched analysis on length, platform, and repository popularity is left to future work.

Two additional correlates reinforce this pattern.
Defect density increases with skill size (Spearman's $\rho = 0.508$, $p < 10^{-300}$): skills over 500 lines average 4.76 defects, compared with 1.48 for skills under 50 lines.
Defects also cluster across categories: only 33.0\% of defective skills are confined to one category, and safety defects co-occur with portability defects at 2.19$\times$ the rate expected by chance.

\subsection{RQ4: What traits characterize reusable, high-quality Agent Skills?}

The previous results identify failure modes.
We next ask what public skills look like when those failure modes are absent.
We identified 2,406 skills from repositories with $\geq$10{,}000 GitHub stars and analyzed the 419 with zero detected defects.
We use these skills as exemplars for qualitative characterization, not as proof that zero detected defects are necessary or sufficient for task success.

These exemplary skills exhibit two distinct styles:
\emph{Minimal-and-precise}: React's \texttt{verify} skill (24 lines) contains only imperative commands and common mistakes.
\emph{Structured-workflow}: Discourse's service-authoring skill (220 lines) defines an eight-phase workflow with gate conditions and audit checklists; every line is actionable.

Five shared traits characterize these exemplary skills:
\begin{enumerate}[leftmargin=*]
    \item \textbf{Trigger-complete descriptions} (G1): every exemplary skill includes ``Use when \ldots''
    \item \textbf{No name-as-heading} (G4): none duplicate their name as an H1 heading.
    \item \textbf{Imperative throughout} (G5): body text consists of directives, not explanations.
    \item \textbf{Project-specific knowledge}: skills encode \emph{local} conventions, not generic programming concepts.
    \item \textbf{No prohibited content} (G8): no install instructions, changelogs, or license text.
\end{enumerate}

We note that traits 1, 2, 3, and 5 are partly definitional under our detectors: a skill with zero detected defects necessarily passes G1, G4, G5, and G8 because those guidelines correspond to the very detectors we use to select the subset.
The genuinely non-tautological observation is trait 4 (project-specific knowledge), which we identified by manual reading of the 419 exemplars: across both the minimal and structured-workflow styles, exemplars consistently encode \emph{local} procedural knowledge (specific to a repository, framework, or service) rather than reproducing tutorial content that an LLM could already generate on demand.
We treat this as a qualitative observation worth following up with an operationalized measure (e.g., embedding distance to LLM-generated baselines for the same description) rather than as a quantified finding.
The common pattern is not merely absence of lint findings: these skills encode local, actionable procedures with clear activation conditions and little reusable-context waste.
This characterization motivates the guideline table below, which combines official structural requirements with empirical and safety-derived constraints.

\subsection{Evidence-Based Authoring Guidelines}

Based on prevalence, platform variation, issue corroboration, and exemplar analysis, we derive twelve guidelines (G1--G12).
Table~\ref{tab:guidelines} maps each guideline to target defects and the percentage of skills affected.

\begin{table*}[htbp]
\caption{Evidence-based skill authoring guidelines, ordered by impact.}
\label{tab:guidelines}
\centering
\small
\begin{tabular}{p{1.8cm}p{6.2cm}p{2.8cm}r}
\toprule
Phase & Guideline & Target Defects & Affected \\
\midrule
\multirow{3}{1.8cm}{\textbf{1. Description}}
& \textbf{G1.} Write $\geq$30 characters: ``\emph{[Verb] [what]}. Use when \emph{[trigger]}.''
& R1.2, R1.3, R1.4
& 52.3\% \\
\cmidrule{2-4}
& \textbf{G2.} Place routing info in \texttt{description}, not body.
& R1.6
& 5.2\% \\
\cmidrule{2-4}
& \textbf{G3.} Keep under 250 characters; front-load the use case.
& R1.5
& 14.3\% \\
\midrule
\multirow{3}{1.8cm}{\textbf{2. Body}}
& \textbf{G4.} Do not repeat \texttt{name} as H1 heading.
& R2.4
& 44.3\% \\
\cmidrule{2-4}
& \textbf{G5.} Write imperative directives, not explanatory prose.
& R2.2, R2.3
& 3.2\% \\
\cmidrule{2-4}
& \textbf{G6.} Keep body under 500 lines.
& R2.1
& 10.4\% \\
\midrule
\textbf{3. Resources}
& \textbf{G7.} Externalize code ($>$60\%) and examples ($>$8 blocks).
& R3.1--R3.3
& 37.0\% \\
\midrule
\textbf{4. Content}
& \textbf{G8.} Remove install instructions, changelogs, licenses, TODOs.
& R4.1--R4.4
& 16.8\% \\
\midrule
\multirow{2}{1.8cm}{\textbf{5. Safety}}
& \textbf{G9.} Remove credentials, user paths, \texttt{--no-verify} flags.
& R5.1, R5.2, R5.6
& 5.8\% \\
\cmidrule{2-4}
& \textbf{G10.} Require confirmation for destructive actions.
& R5.4, R5.5
& 4.7\% \\
\midrule
\textbf{6. Portability}
& \textbf{G11.} No hardcoded models, platform paths, or OS commands.
& R6.1--R6.4
& 5.8\% \\
\midrule
\textbf{7. Identity}
& \textbf{G12.} Do not redefine agent persona or override prior instructions.
& R7.1--R7.3
& 6.8\% \\
\bottomrule
\end{tabular}
\end{table*}

\section{Discussion}
\label{sec:discussion}

\paragraph{The conversation-first provenance hypothesis.}
Why are detected defect rates so high despite a clear specification?
A reusable skill is not merely a saved prompt; it is a routed, progressively disclosed, executable capability.
We propose---and label explicitly as a \emph{hypothesis} rather than a measured finding---that many public skills are instead extracted from problem-solving conversations: a developer solves a task interactively, then saves the conversation as \texttt{SKILL.md}.
This would explain the recurring co-occurrence of routing metadata weakness, name-as-heading duplication, install instructions, hardcoded paths, model names, and safety bypasses, which together resemble what a chat transcript looks like when serialized to disk.
The platform and AI-marker associations are consistent with this hypothesis but do not establish it: the exact authoring path of each public skill is not observable from the file alone, and we did not measure text overlap with chat transcripts or look for explicit ``saved from chat'' markers, both of which would be required to test the hypothesis directly.
We surface it here because it offers a single mechanism that ties together the observed defect cluster and motivates the generation-time interventions in the workflow below; we leave its empirical evaluation to future work.

\paragraph{Specification compliance is necessary but not sufficient.}
Anthropic's official structure provides the minimal contract for reuse: frontmatter descriptions make skills routable, body-size guidance limits context cost, and resource directories enable progressive disclosure.
Our data supports this guidance: spec-aware skills average fewer detected defects, and R1-clean skills perform better in the routing stress test.
At the same time, a syntactically valid skill can still be generic, stale, unsafe, or tied to a local environment; conversely, a cosmetic defect such as repeating the name as an H1 heading is less serious than shipping a live API key.
This distinction motivates our taxonomy: official-spec conformance (R1--R3) is separated from best-practice and operational risks (R4--R7), and our workflow combines spec-aware prompting with linting, repair, and safety gates rather than treating official formatting as sufficient.

\paragraph{Real-world impact: evidence from GitHub issues.}
Our issue analysis classified 303 of 761 relevant issues into our R1--R7 taxonomy.
Routing defects (R1) are both the most prevalent (67.0\%) and most reported (69 issues), followed by body/content issues (75), safety issues (62 plus 108 security-scan disputes), and scope/persona issues (12).
Notable examples include skills that never activate despite matching descriptions, 28 plugin skills totaling 34K tokens causing 4+ minute cold starts, and published skills shipping live API keys.
Of 108 security-scan false-positive appeals on ClawHub, 63 (58.3\%) were triggered by credential-related patterns in legitimate security-auditing skills, confirming that keyword-based scanners cannot distinguish skills that \emph{handle} credentials from those that \emph{leak} them.

\paragraph{Prioritizing defects.}
Defect-free status is not the goal.
Critical defects are routing failures (R1.1--R1.4) and safety violations (R5), because they prevent activation or create vulnerabilities.
Important defects such as body bloat and monolithic structure waste context and hinder maintenance.
Cosmetic defects such as name-as-heading duplication waste tokens but may not break execution.
For practitioners, the priority is eliminating high-impact defects, especially hardcoded credentials, safety bypasses, unguarded destructive actions, and user path leakage.

\begin{figure}[htbp]
\centering
\includegraphics[width=\columnwidth]{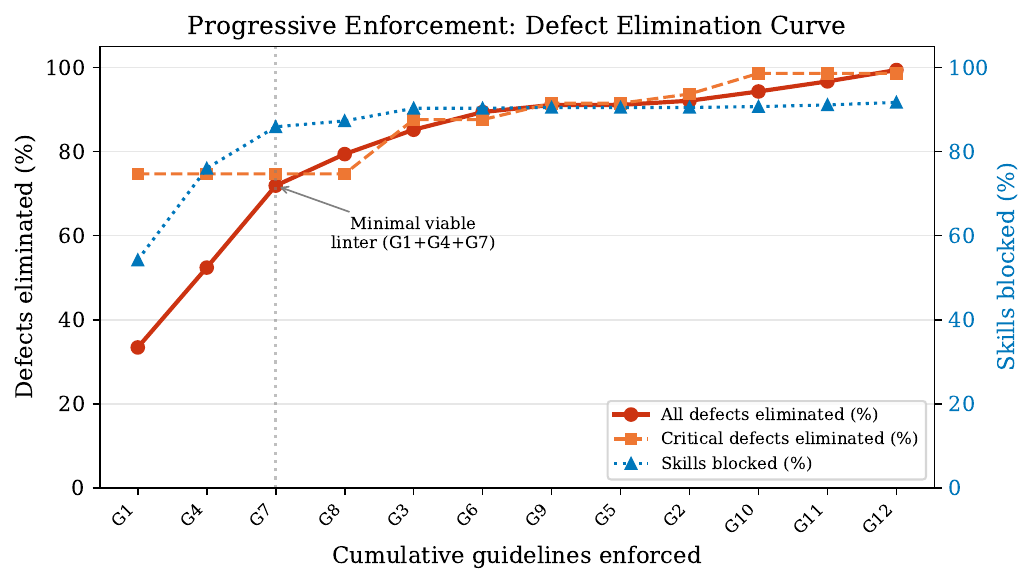}
\caption{Progressive enforcement: cumulative coverage of observed defect instances as guidelines are added in order of marginal defect coverage (not in the phase-ordered sequence used in Table~\ref{tab:guidelines}). The elbow at G1+G4+G7 marks the three-rule lightweight linter.}
\label{fig:progressive}
\end{figure}

\paragraph{Progressive enforcement: quantifying the return on investment.}
We simulated 12 cumulative linter configurations by adding guidelines in order of marginal defect coverage.
Figure~\ref{fig:progressive} shows a clear elbow at three guidelines: G1 catches 33.4\% of observed defect instances, and adding G4 and G7 raises coverage to \textbf{71.9\%} while flagging 85.9\% of skills.
The full suite covers 99.4\% of observed defect instances, while adding G9 and G10 raises critical-defect coverage (routing + safety) from 74.7\% to 98.6\%.

\paragraph{A quality-assured generation workflow.}
Because self-generated skills provide no average benefit in SkillsBench~\cite{li2026skillsbenchbenchmarkingagentskills}, generation workflows need explicit quality gates.
Our findings motivate four stages that correspond to the reuse lifecycle: write routable skills, catch cheap structural defects, repair fixable content, and block unsafe deployment.

\textbf{Stage~1: Spec-aware prompting.}
Specification awareness is the strongest quality correlate we identified (Cliff's $\delta = -0.40$, medium effect): skills whose descriptions follow the ``\emph{[Verb] [what]}. Use when \emph{[trigger]}'' pattern average 1.83 detected defects vs.\ 3.00 for spec-unaware skills.
Generation prompts should include routing requirements (G1--G3), structural constraints (G4--G7), and content prohibitions (G8).

\textbf{Stage~2: Lightweight linting.}
Even spec-aware generation will not eliminate all defects.
The three-rule linter (G1, G4, G7) acts as a fast deterministic gate before repair or deployment.

\textbf{Stage~3: Automated repair.}
Skills that fail linting enter an LLM-based repair loop.
Our repair experiment on 200 defective skills shows that providing the LLM with the detected defect list and relevant guidelines fixes 58.4\% of detected defects while introducing only 0.06 new detected defects per skill.
Structural defects are repaired reliably, but content-reduction defects remain difficult; heavily defective skills should be regenerated rather than repaired.

\textbf{Stage~4: Safety gating.}
In our repair experiment, some safety defects (R5: hardcoded credentials, safety bypasses, prompt injection, unguarded destructive commands) were not repaired reliably, with 0\% fix rate for bypass and injection defects.
For that reason, safety checks should be non-bypassable gates after repair and before deployment, with R5 subcategories guiding context-aware review beyond keyword matching.

\section{Threats to Validity}
\label{sec:threats}

\paragraph{Construct validity.}
Our defect detectors use regex-based heuristics, which may produce false positives or false negatives.
We mitigate implementation risk through mutation-style fixture tests and threshold sensitivity analysis, but these checks do not capture every semantic context in which a skill may be used.
The risk is highest for the two most context-dependent categories.
R5.3 (prompt injection) catches lexical patterns that can appear legitimately inside security-auditing or red-team skills that deliberately discuss injection prompts.
R7.1 (persona redefinition) catches second-person framing such as ``you are a\ldots'' that is appropriate in skills designed to teach an agent a specialized role.
A stratified manual precision audit of these safety-critical detectors is the most important construct-validity check we did not run, and we flag it as the first item of follow-up work; the headline 91.8\% number is robust to it because R5 and R7 contribute only a small share of the all-category count, but per-category R5/R7 prevalence should be read as upper bounds.
The 91.8\% figure also bundles defects of very different severity---cosmetic findings such as name-as-heading duplication (R2.4, 44.3\%) are counted alongside shipped credentials---so we additionally report the Tier~1 89.3\% number throughout and discuss severity prioritization in Section~\ref{sec:discussion} so that the headline is not read as ``92\% of skills are broken.''
Ambiguous credential-handling or persona-based skills may still be misclassified, and downstream task success may depend on context beyond static file structure.
Treating the official specification as the reference also means that legitimate deviations from the recommended format (narrative-style skills, longer-than-recommended bodies that remain actionable) can be labeled as defects when they may work well in practice; the distinction between Tier~1 (spec) and Tier~2 (best practice) is partly motivated by this asymmetry, and we recommend reading defect counts as static-analysis findings rather than as direct evidence of task-time failure.

\paragraph{Internal validity.}
Some detected ``defects'' may be intentional design choices (e.g., persona redefinition for specialized skills).
We separate spec violations from best-practice recommendations and report prevalence rather than normative claims.
The repair experiment uses GPT-4o-mini and a sample of 200 skills; stronger models may perform differently, and the per-category fix rates (notably the 0\% for R5.2 bypass and R5.3 injection) are based on small per-category subsamples that we therefore use only to motivate the safety-gating stage, not to argue precise repair-rate point estimates.
Platform and AI-marker analyses identify ecosystem patterns, but the underlying authoring process is not directly observed.
The cross-platform comparison (Figure~\ref{fig:crossplatform}) further rests on the 41.7\% of skills that file paths allow us to attribute; the remaining 58.3\% land in generic \texttt{skills/} paths and are excluded from per-platform statistics, so platform-level claims should be read as conditional on the attributable subset.

\paragraph{External validity.}
Our dataset is limited to public GitHub repositories and the agentskills.in registry; enterprise skills may differ.
The top 5 repositories contribute 23.5\% of skills; we report both aggregate and per-repository statistics.
Platform attribution via file paths is approximate because generic \texttt{skills/} paths (58.3\%) cannot be attributed to a specific platform.

\paragraph{Functional generalization beyond BM25 and beyond R1.}
The routing stress test isolates the discovery stage of reuse with a BM25 retriever (Section~\ref{sec:routing-stress}); production harnesses use LLM-based selection over the full description set, so the absolute Hit@1 numbers should not be read as production routing performance, and the gap between routing-clean and routing-defective skills may compress under semantic selection.
We also do not run end-to-end task benchmarks for R2--R7; their operational impact is supported only indirectly through GitHub-issue corroboration, our repair experiment, and prior work~\cite{li2026skillsbenchbenchmarkingagentskills,gao2026skillreduceroptimizingllmagent,liu2026agenticskillsworkwild}.
Extending the stress test with an embedding reranker, an LLM selector, and a harness-level benchmark over the remaining defect categories is a primary direction we intend to pursue in follow-up work.

\paragraph{Reliability.}
The specification is evolving; rules grounded in the current version may not apply to future revisions.
We document the spec version and detection logic for reproducibility.

\section{Conclusion}
\label{sec:conclusion}

We studied what keeps public Agent Skills from becoming reusable agent capabilities.
Across 138,133 public \texttt{SKILL.md} files, 89.3\% violate at least one official-specification rule and 91.8\% contain at least one detected reusability defect under our baseline detector, with the finding stable under lenient and strict thresholds.
Routing is the clearest functional bottleneck we isolate: skills with clean routing metadata are retrieved more reliably than those with routing defects from the startup description surface under a BM25 lexical baseline; an LLM-based selector may compress this gap, and we treat lexical retrieval as a lower-bound probe of discovery-stage reuse.

These findings lead to a four-stage quality-assured generation workflow: spec-aware prompting, lightweight linting, automated repair, and safety gating, supported by static analysis, routing stress testing, enforcement simulation, and repair experiments on our dataset.
The workflow addresses the practical reality that self-generated skills currently provide little average benefit by embedding routing, structure, resource organization, and safety checks into generation itself.

\paragraph{Data Availability.}
Our dataset (138,133 skills, CC-BY-4.0) and all analysis scripts are publicly available at \url{https://huggingface.co/datasets/FayeZC/SkillMD-138K}.


\bibliographystyle{ACM-Reference-Format}
\bibliography{references}


\begin{thebibliography}{16}


\ifx \showCODEN    \undefined \def \showCODEN     #1{\unskip}     \fi
\ifx \showDOI      \undefined \def \showDOI       #1{#1}\fi
\ifx \showISBNx    \undefined \def \showISBNx     #1{\unskip}     \fi
\ifx \showISBNxiii \undefined \def \showISBNxiii  #1{\unskip}     \fi
\ifx \showISSN     \undefined \def \showISSN      #1{\unskip}     \fi
\ifx \showLCCN     \undefined \def \showLCCN      #1{\unskip}     \fi
\ifx \shownote     \undefined \def \shownote      #1{#1}          \fi
\ifx \showarticletitle \undefined \def \showarticletitle #1{#1}   \fi
\ifx \showURL      \undefined \def \showURL       {\relax}        \fi
\providecommand\bibfield[2]{#2}
\providecommand\bibinfo[2]{#2}
\providecommand\natexlab[1]{#1}
\providecommand\showeprint[2][]{arXiv:#2}

\bibitem[{agent-sh}(2026)]%
        {agnix}
\bibfield{author}{\bibinfo{person}{{agent-sh}}.}
  \bibinfo{year}{2026}\natexlab{}.
\newblock \bibinfo{title}{agnix: The missing linter and {LSP} for {AI} coding
  assistants}.
\newblock
\newblock
\urldef\tempurl%
\url{https://github.com/agent-sh/agnix}
\showURL{%
Retrieved May 4, 2026 from \tempurl}


\bibitem[{Agent Skills}({[n.\,d.]})]%
        {agentskills-spec}
\bibfield{author}{\bibinfo{person}{{Agent Skills}}.}
  \bibinfo{year}{[n.\,d.]}\natexlab{}.
\newblock \bibinfo{title}{Specification - Agent Skills}.
\newblock
\newblock
\urldef\tempurl%
\url{https://agentskills.io/specification}
\showURL{%
Retrieved May 4, 2026 from \tempurl}


\bibitem[Anthropic({[n.\,d.]})]%
        {claudecode-skills}
\bibfield{author}{\bibinfo{person}{Anthropic}.}
  \bibinfo{year}{[n.\,d.]}\natexlab{}.
\newblock \bibinfo{title}{Extend {Claude} with skills}.
\newblock
\newblock
\urldef\tempurl%
\url{https://code.claude.com/docs/en/skills}
\showURL{%
Retrieved May 4, 2026 from \tempurl}


\bibitem[Gao et~al\mbox{.}(2026)]%
        {gao2026skillreduceroptimizingllmagent}
\bibfield{author}{\bibinfo{person}{Yudong Gao}, \bibinfo{person}{Zongjie Li},
  \bibinfo{person}{Yuanyuanyuan}, \bibinfo{person}{Zimo Ji},
  \bibinfo{person}{Pingchuan Ma}, {and} \bibinfo{person}{Shuai Wang}.}
  \bibinfo{year}{2026}\natexlab{}.
\newblock \bibinfo{title}{SkillReducer: Optimizing LLM Agent Skills for Token
  Efficiency}.
\newblock
\newblock
\urldef\tempurl%
\url{https://doi.org/10.48550/arXiv.2603.29919}
\showDOI{\tempurl}
\showeprint[arxiv]{2603.29919}~[cs.SE]


\bibitem[Geng et~al\mbox{.}(2026)]%
        {geng2025control}
\bibfield{author}{\bibinfo{person}{Yilin Geng}, \bibinfo{person}{Haonan Li},
  \bibinfo{person}{Honglin Mu}, \bibinfo{person}{Xudong Han},
  \bibinfo{person}{Timothy Baldwin}, \bibinfo{person}{Omri Abend},
  \bibinfo{person}{Eduard Hovy}, {and} \bibinfo{person}{Lea Frermann}.}
  \bibinfo{year}{2026}\natexlab{}.
\newblock \showarticletitle{Control Illusion: The Failure of Instruction
  Hierarchies in Large Language Models}. In
  \bibinfo{booktitle}{\emph{Proceedings of the {AAAI} Conference on Artificial
  Intelligence}}, Vol.~\bibinfo{volume}{40}. \bibinfo{pages}{30816--30824}.
\newblock
\urldef\tempurl%
\url{https://doi.org/10.1609/aaai.v40i36.40339}
\showDOI{\tempurl}


\bibitem[Greshake et~al\mbox{.}(2023)]%
        {greshake2023indirect}
\bibfield{author}{\bibinfo{person}{Kai Greshake}, \bibinfo{person}{Sahar
  Abdelnabi}, \bibinfo{person}{Shailesh Mishra}, \bibinfo{person}{Christoph
  Endres}, \bibinfo{person}{Thorsten Holz}, {and} \bibinfo{person}{Mario
  Fritz}.} \bibinfo{year}{2023}\natexlab{}.
\newblock \showarticletitle{Not What You've Signed Up For: Compromising
  Real-World {LLM}-Integrated Applications with Indirect Prompt Injection}. In
  \bibinfo{booktitle}{\emph{Proceedings of the 16th ACM Workshop on Artificial
  Intelligence and Security}}. \bibinfo{pages}{79--90}.
\newblock
\urldef\tempurl%
\url{https://doi.org/10.1145/3605764.3623985}
\showDOI{\tempurl}


\bibitem[Guo et~al\mbox{.}(2026)]%
        {guo2026ihchallengetrainingdatasetimprove}
\bibfield{author}{\bibinfo{person}{Chuan Guo}, \bibinfo{person}{Juan
  Felipe~Ceron Uribe}, \bibinfo{person}{Sicheng Zhu},
  \bibinfo{person}{Christopher~A. Choquette-Choo}, \bibinfo{person}{Steph Lin},
  \bibinfo{person}{Nikhil Kandpal}, \bibinfo{person}{Milad Nasr},
  \bibinfo{person}{Rai}, \bibinfo{person}{Sam Toyer}, \bibinfo{person}{Miles
  Wang}, \bibinfo{person}{Yaodong Yu}, \bibinfo{person}{Alex Beutel}, {and}
  \bibinfo{person}{Kai Xiao}.} \bibinfo{year}{2026}\natexlab{}.
\newblock \bibinfo{title}{IH-Challenge: A Training Dataset to Improve
  Instruction Hierarchy on Frontier LLMs}.
\newblock
\newblock
\urldef\tempurl%
\url{https://doi.org/10.48550/arXiv.2603.10521}
\showDOI{\tempurl}
\showeprint[arxiv]{2603.10521}~[cs.AI]


\bibitem[{ISO/IEC}(2023)]%
        {iso25010}
\bibfield{author}{\bibinfo{person}{{ISO/IEC}}.}
  \bibinfo{year}{2023}\natexlab{}.
\newblock \bibinfo{title}{{ISO/IEC 25010}:2023 --- Systems and Software Quality
  Requirements and Evaluation ({SQuaRE}) --- Product Quality Model}.
\newblock
\newblock
\urldef\tempurl%
\url{https://www.iso.org/standard/78176.html}
\showURL{%
\tempurl}


\bibitem[Li et~al\mbox{.}(2026a)]%
        {li2026skillsbenchbenchmarkingagentskills}
\bibfield{author}{\bibinfo{person}{Xiangyi Li}, \bibinfo{person}{Wenbo Chen},
  \bibinfo{person}{Yimin Liu}, \bibinfo{person}{Shenghan Zheng},
  \bibinfo{person}{Xiaokun Chen}, \bibinfo{person}{Yifeng He},
  \bibinfo{person}{Yubo Li}, \bibinfo{person}{Bingran You},
  \bibinfo{person}{Haotian Shen}, \bibinfo{person}{Jiankai Sun},
  \bibinfo{person}{Shuyi Wang}, \bibinfo{person}{Binxu Li},
  \bibinfo{person}{Qunhong Zeng}, \bibinfo{person}{Di Wang},
  \bibinfo{person}{Xuandong Zhao}, \bibinfo{person}{Yuanli Wang},
  \bibinfo{person}{Roey~Ben Chaim}, \bibinfo{person}{Zonglin Di},
  \bibinfo{person}{Yipeng Gao}, \bibinfo{person}{Junwei He},
  \bibinfo{person}{Yizhuo He}, \bibinfo{person}{Liqiang Jing},
  \bibinfo{person}{Luyang Kong}, \bibinfo{person}{Xin Lan},
  \bibinfo{person}{Jiachen Li}, \bibinfo{person}{Songlin Li},
  \bibinfo{person}{Yijiang Li}, \bibinfo{person}{Yueqian Lin},
  \bibinfo{person}{Xinyi Liu}, \bibinfo{person}{Xuanqing Liu},
  \bibinfo{person}{Haoran Lyu}, \bibinfo{person}{Ze Ma}, \bibinfo{person}{Bowei
  Wang}, \bibinfo{person}{Runhui Wang}, \bibinfo{person}{Tianyu Wang},
  \bibinfo{person}{Wengao Ye}, \bibinfo{person}{Yue Zhang},
  \bibinfo{person}{Hanwen Xing}, \bibinfo{person}{Yiqi Xue},
  \bibinfo{person}{Steven Dillmann}, {and} \bibinfo{person}{Han chung Lee}.}
  \bibinfo{year}{2026}\natexlab{a}.
\newblock \bibinfo{title}{SkillsBench: Benchmarking How Well Agent Skills Work
  Across Diverse Tasks}.
\newblock
\newblock
\urldef\tempurl%
\url{https://doi.org/10.48550/arXiv.2602.12670}
\showDOI{\tempurl}
\showeprint[arxiv]{2602.12670}~[cs.AI]


\bibitem[Li et~al\mbox{.}(2026b)]%
        {li2026clawsbenchevaluatingcapabilitysafety}
\bibfield{author}{\bibinfo{person}{Xiangyi Li}, \bibinfo{person}{Kyoung~Whan
  Choe}, \bibinfo{person}{Yimin Liu}, \bibinfo{person}{Xiaokun Chen},
  \bibinfo{person}{Chujun Tao}, \bibinfo{person}{Bingran You},
  \bibinfo{person}{Wenbo Chen}, \bibinfo{person}{Zonglin Di},
  \bibinfo{person}{Jiankai Sun}, \bibinfo{person}{Shenghan Zheng},
  \bibinfo{person}{Jiajun Bao}, \bibinfo{person}{Yuanli Wang},
  \bibinfo{person}{Weixiang Yan}, \bibinfo{person}{Yiyuan Li}, {and}
  \bibinfo{person}{Han chung Lee}.} \bibinfo{year}{2026}\natexlab{b}.
\newblock \bibinfo{title}{ClawsBench: Evaluating Capability and Safety of LLM
  Productivity Agents in Simulated Workspaces}.
\newblock
\newblock
\urldef\tempurl%
\url{https://doi.org/10.48550/arXiv.2604.05172}
\showDOI{\tempurl}
\showeprint[arxiv]{2604.05172}~[cs.AI]


\bibitem[Liu et~al\mbox{.}(2026)]%
        {liu2026agenticskillsworkwild}
\bibfield{author}{\bibinfo{person}{Yujian Liu}, \bibinfo{person}{Jiabao Ji},
  \bibinfo{person}{Li An}, \bibinfo{person}{Tommi Jaakkola},
  \bibinfo{person}{Yang Zhang}, {and} \bibinfo{person}{Shiyu Chang}.}
  \bibinfo{year}{2026}\natexlab{}.
\newblock \bibinfo{title}{How Well Do Agentic Skills Work in the Wild:
  Benchmarking LLM Skill Usage in Realistic Settings}.
\newblock
\newblock
\urldef\tempurl%
\url{https://doi.org/10.48550/arXiv.2604.04323}
\showDOI{\tempurl}
\showeprint[arxiv]{2604.04323}~[cs.CL]


\bibitem[{MITRE Corporation}(2026)]%
        {cwe798}
\bibfield{author}{\bibinfo{person}{{MITRE Corporation}}.}
  \bibinfo{year}{2026}\natexlab{}.
\newblock \bibinfo{title}{{CWE-798}: Use of Hard-coded Credentials}.
\newblock
\newblock
\urldef\tempurl%
\url{https://cwe.mitre.org/data/definitions/798.html}
\showURL{%
Retrieved May 4, 2026 from \tempurl}


\bibitem[{OWASP Gen AI Security Project}(2025)]%
        {owasp2025}
\bibfield{author}{\bibinfo{person}{{OWASP Gen AI Security Project}}.}
  \bibinfo{year}{2025}\natexlab{}.
\newblock \bibinfo{title}{2025 Top 10 Risk \& Mitigations for {LLMs} and Gen
  {AI} Apps}.
\newblock
\newblock
\urldef\tempurl%
\url{https://genai.owasp.org/llm-top-10/}
\showURL{%
Retrieved May 4, 2026 from \tempurl}


\bibitem[Schmotz et~al\mbox{.}(2026)]%
        {schmotz2026skillinjectmeasuringagentvulnerability}
\bibfield{author}{\bibinfo{person}{David Schmotz}, \bibinfo{person}{Luca
  Beurer-Kellner}, \bibinfo{person}{Sahar Abdelnabi}, {and}
  \bibinfo{person}{Maksym Andriushchenko}.} \bibinfo{year}{2026}\natexlab{}.
\newblock \bibinfo{title}{Skill-Inject: Measuring Agent Vulnerability to Skill
  File Attacks}.
\newblock
\newblock
\urldef\tempurl%
\url{https://doi.org/10.48550/arXiv.2602.20156}
\showDOI{\tempurl}
\showeprint[arxiv]{2602.20156}~[cs.CR]


\bibitem[Wallace et~al\mbox{.}(2024)]%
        {wallace2024instructionhierarchytrainingllms}
\bibfield{author}{\bibinfo{person}{Eric Wallace}, \bibinfo{person}{Kai Xiao},
  \bibinfo{person}{Reimar Leike}, \bibinfo{person}{Lilian Weng},
  \bibinfo{person}{Johannes Heidecke}, {and} \bibinfo{person}{Alex Beutel}.}
  \bibinfo{year}{2024}\natexlab{}.
\newblock \bibinfo{title}{The Instruction Hierarchy: Training LLMs to
  Prioritize Privileged Instructions}.
\newblock
\newblock
\urldef\tempurl%
\url{https://doi.org/10.48550/arXiv.2404.13208}
\showDOI{\tempurl}
\showeprint[arxiv]{2404.13208}~[cs.CR]


\bibitem[Wang et~al\mbox{.}(2026)]%
        {wang2026elementarydearwatsondetecting}
\bibfield{author}{\bibinfo{person}{Shenao Wang}, \bibinfo{person}{Junjie He},
  \bibinfo{person}{Yanjie Zhao}, \bibinfo{person}{Yayi Wang},
  \bibinfo{person}{Kan Yu}, {and} \bibinfo{person}{Haoyu Wang}.}
  \bibinfo{year}{2026}\natexlab{}.
\newblock \bibinfo{title}{"Elementary, My Dear Watson." Detecting Malicious
  Skills via Neuro-Symbolic Reasoning across Heterogeneous Artifacts}.
\newblock
\newblock
\urldef\tempurl%
\url{https://doi.org/10.48550/arXiv.2603.27204}
\showDOI{\tempurl}
\showeprint[arxiv]{2603.27204}~[cs.CR]


\end{thebibliography}

\end{document}